\documentclass{article}
\usepackage{spconf,amsmath,graphicx,hyperref,booktabs,amssymb,xcolor, tabularx}

\newcommand{\pgraph}[1]{\par\smallskip\noindent\textbf{#1}\hskip 0.6em\relax}

\title{Is Spectral Predictability Enough? Testing $\Omega$-Matching for
  Synthetic Pretraining under Distribution Shift}

\name{Edric Castel C.~Hao, Juan Raphael S.~Se\~{n}a, Neon S.~De Jesus}
\address{Analog Devices, Inc.}

\begin{document}

\maketitle

\begin{abstract}
Synthetic pretraining under distribution shift requires deciding what
synthetic data to use. A natural criterion is to match the spectral
predictability $\Omega$ of the source to the target, but is $\Omega$
sufficient? We test this hypothesis, which we call
\emph{predictability-aware} pretraining, across three architectures
(LSTM, PatchTST, TCN) and three seeds. Under ideal synthetic-to-synthetic
conditions, $\Omega$-matched pretraining can outperform even an
oracle baseline ($+46.8\%$ MSE reduction, LSTM), though the benefit is
already architecture-dependent. On the real-world ETTm1
benchmark, however, this relationship does not hold: sine-plus-noise
pretraining at matched $\Omega$ yields only a marginal mean improvement
($+2.3\%$ averaged over all three architectures, up to $+65.5\%$ on
individual channels), whereas $\Omega$-matched chaotic ODE trajectories
systematically \emph{degrade} performance ($-13.2\%$). The two sources
agree on the direction of transfer for only $53\%$ of configurations
across all three architectures (Pearson $r=-0.02$), indicating
that $\Omega$ carries little information about which source will transfer
beneficially. We further observe an architecture-dependent asymmetry in
the response to the direction of the $\Omega$ gap. We conclude that
$\Omega$ matching can help in controlled settings but is not sufficient to
predict transfer across sources, and we discuss additional signal
properties that a source-selection criterion may need to consider.
\end{abstract}


\section{Introduction}

Variable renewable sources (solar PV and wind) lower the
\emph{spectral predictability} of grid load signals, making multi-step
forecasting harder~\cite{renewables_shift}. Predictability can be
quantified via $\Omega \in [0,1]$, where high $\Omega$ indicates a strongly periodic signal and low $\Omega$ is chaotic-like~\cite{spectral_pred_paper}. This
raises a concrete, practical question for anyone building forecasters for
shifting regimes:

\begin{quote}
\emph{Can $\Omega$ matching serve as a reliable criterion for selecting
synthetic pretraining data under distribution shift?}
\end{quote}

This criterion enables a practical data-generation strategy. If $\Omega$ controls transfer
quality, one can generate unlimited synthetic pretraining data calibrated
to any target $\Omega$, circumventing the scarcity of real domain data.
The approach is an instance of transfer
learning~\cite{transferlearning_survey} and is related to recent
time-series foundation models that rely on large-scale
(self-)pretraining~\cite{patchformer,droppatch}. We refer to this
hypothesis as \emph{predictability-aware} (\texttt{pred\_aware})
pretraining: synthetic source data are generated with $\Omega$ matched to
the target regime, and the model is subsequently fine-tuned on the limited
available target data.

Rather than proposing \texttt{pred\_aware} as a solution, we treat it as a
hypothesis to be tested. We first establish, under fully controlled
synthetic-to-synthetic conditions, that $\Omega$ matching behaves as
hypothesized. We then
assess whether this behavior persists on real data by pretraining on two
markedly different $\Omega$-matched synthetic sources (sine-plus-noise and
chaotic ODE trajectories) and fine-tuning on the ETTm1 benchmark. Contrary
to expectation, the two sources, matched to the same target
$\Omega$, produce sharply divergent and near-uncorrelated outcomes. This
constitutes a \emph{negative result}: $\Omega$ alone is not a reliable
criterion for source selection. We conclude by discussing candidate
factors that such a criterion may need to incorporate.

\section{Background and Method}

\pgraph{Spectral predictability $\Omega$.}
Following~\cite{spectral_pred_paper}, we define
$\Omega = 1 - H_{\mathrm{norm}}$, where $H_{\mathrm{norm}}$ is the
normalized spectral entropy of the signal's power spectral density (PSD),
so $\Omega=1$ for periodic signals and $\Omega=0$ for white
noise~\cite{spectral_entropy}. $\Omega$ is one of several forecastability
measures~\cite{forecastability,beyond_ranking} that summarize how hard a
series is to predict. We write $\omega_{\mathrm{pre}}$ for the
pretraining $\Omega$ and $\omega_{\mathrm{tgt}}$ for the target, with gap
$\Delta\omega = \omega_{\mathrm{pre}} - \omega_{\mathrm{tgt}}$.

\pgraph{Estimating the target $\Omega$.}
The target $\Omega$ is estimated from the PSD of the first $T$ rows of
ETTm1 (i.e., the fine-tuning prefix a practitioner would actually
have), using the spectral-entropy formula above. Computed this way, the
per-channel target $\Omega$ of ETTm1 ranges from $0.36$ to $0.76$ across
the seven channels, which is the range \texttt{pred\_aware} attempts to
match.

\pgraph{Conditions.}
We compare four conditions, each testing a specific hypothesis:
\texttt{no\_aug} (no pretraining; the reference point),
\texttt{pred\_aware\_aug} (synthetic pretraining at $\Omega$ matched to
$\omega_{\mathrm{tgt}}$; the hypothesis under test),
\texttt{random\_\allowbreak synthetic\_\allowbreak aug} ($\Omega$ drawn
uniformly from $[0,1]$;
an $\Omega$-agnostic synthetic baseline that isolates the value of
\emph{matching}), and \texttt{oracle} (pretraining on real target-domain
data; an upper bound, available only in the synthetic-to-synthetic study).

\pgraph{Fine-tuning.}
All ETTm1 experiments use a two-stage procedure. \emph{Stage~1}: pretrain
the model on synthetic data generated at the matched $\Omega$.
\emph{Stage~2}: fine-tune the pretrained model on the first $T$ rows of
ETTm1 using identical hyperparameters across conditions. Evaluation is on
the held-out test window, ETTm1 rows $46{,}080$--$69{,}679$, which never
overlaps the fine-tuning prefix.

\pgraph{Sine-plus-noise generation.}
Synthetic sequences are
$x(t) = \sum_{k=1}^{K} A_k \sin(2\pi f_k t + \phi_k) + \sigma\,\varepsilon(t)$
with $\varepsilon(t) \sim \mathcal{N}(0,1)$. The noise level $\sigma$
controls $\Omega$ by adjusting the balance between harmonic and broadband
spectral energy.

\pgraph{Dysts ODE generation.}
The second source draws trajectories from the \texttt{dysts}
library~\cite{dysts} of chaotic dynamical systems. For each target
$\omega$ and sequence length, we select the system whose empirical
$\Omega$ most closely matches the target and integrate it to produce the
pretraining sequence. The selected systems span diverse dynamics (e.g.,
\texttt{ExcitableCell}, \texttt{GuckenheimerHolmes}, \texttt{SprottH},
\texttt{StickSlipOscillator}), with achieved $\Omega$ within $\pm0.01$ of
the target except at $\omega=0.9$. Crucially, these trajectories are
matched to the target only in $\Omega$. Their local waveform structure is
otherwise unconstrained, which is central to the source comparison in
Sec.~\ref{sec:compare}.

\pgraph{Architectures.}
We evaluate three families:
LSTM~\cite{hochreiter1997} (recurrent);
PatchTST~\cite{patchtst} (patch-based attention); and
TCN (Temporal Convolutional Network)~\cite{tcn}, a convolutional
architecture with dilated causal convolutions
(kernel\_size=2, dilations=[1,2,4,8,16]).

\pgraph{Experimental setup.}
We evaluate all three models on ETTm1, a standard long-horizon
forecasting benchmark~\cite{autoformer,informer}, with dataset sizes
$T \in \{2000, 10000\}$, all seven channels, and pretraining
$\omega \in \{0.3, 0.5, 0.7, 0.9\}$, over three seeds
($123, 456, 789$). All reported values are seed-averaged.
The $\omega=0.9$ target was not fully achievable: sine-plus-noise reached
$\Omega\approx0.80$--$0.84$ even at zero noise (spectral leakage from
finite sequence length), and the dysts sources reached
$\Omega\approx0.86$--$0.89$. All other targets were achieved within
$\pm0.01$. Results labeled $\omega=0.9$ reflect these achieved values.

\section{Results}

\subsection{Verification under Ideal Conditions (Synthetic $\to$ Synthetic)}

As a controlled verification, we train and test on synthetic data with
known $\Omega$, isolating the $\Omega$ gap from all other confounds and
thereby providing the most favorable setting for the hypothesis. At
$\Delta\omega = +0.4$ (the largest gap tested; down-gap
$\omega_{\mathrm{pre}}0.7\!\to\!\omega_{\mathrm{tgt}}0.3$),
\texttt{pred\_aware} exceeds even the oracle baseline for all three
architectures, most strongly for LSTM ($\sim$31\,pp; Table~\ref{tab:study_a}).
The advantage over the $\Omega$-agnostic \texttt{random\_synthetic}
baseline is, however, small and architecture-dependent: the two are
comparable for LSTM ($+46.8\%$ vs.\ $+46.7\%$) and PatchTST ($+1.3\%$ vs.\
$-3.4\%$), while for TCN \texttt{random\_synthetic} is the stronger of the
two ($+6.4\%$ vs.\ $-2.5\%$). Much of the reduction over \texttt{no\_aug}
at this gap is thus attributable to synthetic pretraining in general
rather than to $\Omega$-matching alone. The sharper, direction-specific
$\Omega$ effect appears in Table~\ref{tab:directional}. We regard this as a
verification step, not the principal result.

\begin{table}[t]
\centering
\caption{Verification (synthetic $\to$ synthetic). MSE reduction vs.\
  \texttt{no\_aug} at $\Delta\omega = +0.4$ (down-gap
  $\omega_{\mathrm{pre}}0.7\!\to\!\omega_{\mathrm{tgt}}0.3$), mean over
  dataset sizes. Positive $=$ improvement; bold $=$ best per model.}
\label{tab:study_a}
\small
\begin{tabular}{lccc}
\toprule
Strategy   & LSTM    & PatchTST & TCN \\
\midrule
random\_synthetic   & $+$46.7\%  & $-$3.4\% & \textbf{$+$6.4\%} \\
oracle              & $+$16.0\% & $-$10.8\% & $-$6.4\% \\
pred\_aware         & \textbf{$+$46.8\%} & \textbf{$+$1.3\%} & $-$2.5\% \\
\bottomrule
\end{tabular}
\end{table}

Even in this ideal setting, the benefit depends on the
\emph{direction} of the gap, and does so differently across architectures
(Table~\ref{tab:directional}). LSTM benefits strongly from down-gap
pretraining ($+46.8\%$, pretraining more predictable than target) but is
harmed by up-gap ($-65.7\%$). PatchTST shows the opposite, gaining from
up-gap ($+86.4\%$) but almost nothing from down-gap ($+1.3\%$). TCN is
near-neutral in both directions ($-2.5\%$ down, $-1.7\%$ up). We return
to this asymmetry in Sec.~\ref{sec:disc}.

\begin{table}[t]
\centering
\caption{Directional asymmetry at $|\Delta\omega|=0.4$. MSE reduction
  (\%) vs.\ \texttt{no\_aug}. Positive $=$ improvement.}
\label{tab:directional}
\small
\begin{tabularx}{\linewidth}{lXXX}
\toprule
Direction & LSTM & PatchTST & TCN \\
\midrule
Down-gap ($0.7\to0.3$) & \textbf{$+46.8\%$} & $+1.3\%$ & $-2.5\%$ \\
Up-gap ($0.3\to0.7$) & $-65.7\%$ & \textbf{$+86.4\%$} & $-1.7\%$ \\
\bottomrule
\end{tabularx}
\end{table}

\subsection{Cross-Domain Transfer: Sine-Plus-Noise $\to$ ETTm1}
\label{sec:b1}

On ETTm1, the pronounced effect observed under ideal conditions largely
disappears. Averaged over all three architectures, mean improvement from
$\Omega$-matched sine-plus-noise pretraining is only $+2.3\%$ over
\texttt{no\_aug}, with large per-channel variance (Tables~\ref{tab:b1_lstm}--\ref{tab:b1_tcn}): the best single
configuration reaches $+65.5\%$ (PatchTST, $T{=}10\text{k}$, HUFL,
$\omega{=}0.9$) while others degrade. Crucially, the $\Omega$ gap no longer
predicts the benefit. Correlating the gap $\Delta\omega$ against each
configuration's normalized MSE tests whether a larger predictability gap systematically
brings better or worse transfer. We find that it does not. The
association is weak for every architecture (LSTM $r{=}0.01$, PatchTST
$r{=}{-}0.03$, TCN $r{=}0.14$; pooled $r{=}0.02$), and even the strongest
case, TCN, explains under $2\%$ of the outcome variance, so the size of the
$\Omega$ gap has essentially no bearing on how well the source transfers. Per-model means (Table~\ref{tab:summary}) show
LSTM and PatchTST improving mildly, more so at $T{=}10\text{k}$, whereas
TCN is flat-to-slightly-harmful ($-3.3\%$ at $T{=}2$k, $-1.2\%$ at
$T{=}10$k; $-2.3\%$ overall): its per-channel table (Table~\ref{tab:b1_tcn}) is dominated by
values $>1.0$, with improvements confined to a few channels. This
indicates that architecture, rather than $\Omega$, is the primary
determinant of the outcome.

\begin{table}[t]
\centering
\caption{LSTM normalized MSE under \textbf{sine-plus-noise} pretraining
(\texttt{pred\_aware}\,/\,\texttt{no\_aug}, $<1.0 =$ improvement, bold)
  by channel and pretrain $\omega$ (seed-averaged).}
\label{tab:b1_lstm}
\scriptsize
\setlength{\tabcolsep}{2.5pt}
\begin{tabular}{l cccc cccc}
\toprule
 & \multicolumn{4}{c}{$T$=2000} & \multicolumn{4}{c}{$T$=10000} \\
\cmidrule(lr){2-5} \cmidrule(lr){6-9}
Ch. & $\omega$.3 & $\omega$.5 & $\omega$.7 & $\omega$.9
        & $\omega$.3 & $\omega$.5 & $\omega$.7 & $\omega$.9 \\
\midrule
HUFL & 1.059 & 1.247 & \textbf{0.988} & 1.155
     & 1.096 & \textbf{0.947} & \textbf{0.819} & \textbf{0.762} \\
HULL & \textbf{0.977} & 1.050 & 1.047 & 1.012
     & \textbf{0.905} & \textbf{0.924} & \textbf{0.930} & \textbf{0.991} \\
LUFL & 1.088 & 1.004 & \textbf{0.930} & \textbf{0.915}
     & \textbf{0.982} & 1.005 & 1.001 & 1.012 \\
LULL & \textbf{0.896} & \textbf{0.991} & 1.010 & \textbf{0.912}
     & \textbf{0.919} & \textbf{0.884} & \textbf{0.858} & \textbf{0.999} \\
MUFL & \textbf{0.595} & 1.058 & \textbf{0.912} & 1.398
     & \textbf{0.938} & 1.003 & \textbf{0.531} & \textbf{0.637} \\
MULL & \textbf{0.922} & 1.048 & 1.053 & \textbf{0.979}
     & \textbf{0.952} & 1.006 & 1.002 & \textbf{0.948} \\
OT   & \textbf{0.953} & 1.000 & 1.014 & \textbf{0.926}
     & 1.000 & \textbf{0.978} & 1.023 & \textbf{0.977} \\
\bottomrule
\end{tabular}
\end{table}

\begin{table}[t]
\centering
\caption{PatchTST normalized MSE under \textbf{sine-plus-noise} pretraining
(\texttt{pred\_aware}\,/\,\texttt{no\_aug}, $<1.0 =$ improvement, bold)
  by channel and pretrain $\omega$ (seed-averaged).}
\label{tab:b1_pt}
\scriptsize
\setlength{\tabcolsep}{2.5pt}
\begin{tabular}{l cccc cccc}
\toprule
 & \multicolumn{4}{c}{$T$=2000} & \multicolumn{4}{c}{$T$=10000} \\
\cmidrule(lr){2-5} \cmidrule(lr){6-9}
Ch. & $\omega$.3 & $\omega$.5 & $\omega$.7 & $\omega$.9
        & $\omega$.3 & $\omega$.5 & $\omega$.7 & $\omega$.9 \\
\midrule
HUFL & \textbf{0.978} & \textbf{0.839} & \textbf{0.844} & \textbf{0.753}
     & \textbf{0.813} & \textbf{0.799} & \textbf{0.697} & \textbf{0.345} \\
HULL & 1.013 & \textbf{0.978} & 1.019 & \textbf{0.945}
     & \textbf{0.936} & \textbf{0.920} & \textbf{0.916} & \textbf{0.997} \\
LUFL & \textbf{0.992} & 1.066 & 1.034 & \textbf{0.994}
     & \textbf{0.915} & \textbf{0.923} & \textbf{0.883} & \textbf{0.920} \\
LULL & \textbf{0.981} & \textbf{0.981} & \textbf{0.942} & \textbf{0.951}
     & \textbf{0.919} & \textbf{0.946} & 1.004 & 1.020 \\
MUFL & \textbf{0.749} & \textbf{0.940} & \textbf{0.835} & \textbf{0.995}
     & \textbf{0.914} & \textbf{0.748} & 1.341 & \textbf{0.891} \\
MULL & 1.040 & 1.015 & 1.017 & 1.024
     & \textbf{0.971} & 1.025 & \textbf{0.974} & \textbf{0.995} \\
OT   & 1.010 & \textbf{0.999} & \textbf{0.991} & \textbf{0.979}
     & \textbf{0.991} & \textbf{0.979} & \textbf{0.986} & \textbf{0.983} \\
\bottomrule
\end{tabular}
\end{table}

\begin{table}[t]
\centering
\caption{TCN normalized MSE under \textbf{sine-plus-noise} pretraining
(\texttt{pred\_aware}\,/\,\texttt{no\_aug}, $<1.0 =$ improvement, bold)
  by channel and pretrain $\omega$ (seed-averaged).}
\label{tab:b1_tcn}
\scriptsize
\setlength{\tabcolsep}{2.5pt}
\begin{tabular}{l cccc cccc}
\toprule
 & \multicolumn{4}{c}{$T$=2000} & \multicolumn{4}{c}{$T$=10000} \\
\cmidrule(lr){2-5} \cmidrule(lr){6-9}
Ch. & $\omega$.3 & $\omega$.5 & $\omega$.7 & $\omega$.9
        & $\omega$.3 & $\omega$.5 & $\omega$.7 & $\omega$.9 \\
\midrule
HUFL & \textbf{0.990} & 1.126 & 1.240 & \textbf{0.868}
     & 1.027 & \textbf{0.832} & \textbf{0.949} & 1.020 \\
HULL & 1.067 & 1.037 & 1.008 & 1.016
     & 1.054 & \textbf{0.986} & 1.042 & 1.062 \\
LUFL & \textbf{0.993} & 1.037 & 1.058 & 1.082
     & 1.039 & 1.047 & 1.028 & 1.057 \\
LULL & 1.067 & 1.025 & 1.040 & 1.059
     & 1.017 & 1.005 & 1.009 & 1.035 \\
MUFL & \textbf{0.960} & 1.149 & 1.068 & \textbf{0.983}
     & \textbf{0.871} & 1.088 & 1.011 & 1.044 \\
MULL & \textbf{0.992} & 1.001 & 1.001 & 1.022
     & 1.043 & 1.004 & 1.035 & 1.041 \\
OT   & 1.021 & 1.014 & 1.018 & \textbf{0.992}
     & \textbf{0.996} & 1.005 & \textbf{0.997} & \textbf{0.984} \\
\bottomrule
\end{tabular}
\end{table}

\begin{table}[t]
\centering
\caption{Mean improvement (\%) vs.\ \texttt{no\_aug} by model and $T$,
  seed-averaged. Positive $=$ improvement. B1: sine-plus-noise.
  B2: dysts ODE.}
\label{tab:summary}
\small
\begin{tabular}{l cc c cc}
\toprule
 & \multicolumn{2}{c}{B1 (sine+noise)} & & \multicolumn{2}{c}{B2 (dysts ODE)} \\
\cmidrule(lr){2-3} \cmidrule(lr){5-6}
Model & $T$=2000 & $T$=10000 & & $T$=2000 & $T$=10000 \\
\midrule
LSTM     & $-$0.5 & $+$7.0 & & $-$2.2  & $-$18.8 \\
PatchTST & $+$3.9 & $+$8.0 & & $-$12.6 & $-$16.7 \\
TCN      & $-$3.3 & $-$1.2 & & $-$7.6  & $-$21.5 \\
\bottomrule
\end{tabular}
\end{table}

\subsection{Cross-Domain Transfer: Dysts ODE $\to$ ETTm1}
\label{sec:b2}

If $\Omega$ were the operative variable, substituting a different
synthetic source matched to the \emph{same} target $\Omega$ would preserve
the outcome, but instead it reverses it. Replacing sine-plus-noise with
$\Omega$-matched chaotic ODE trajectories from
\texttt{dysts}~\cite{dysts} converts the marginal gain into systematic
degradation: mean improvement is $\mathbf{-13.2\%}$. Every model/size cell is negative
(Table~\ref{tab:summary}), with extreme worst cases such
as $-243.6\%$ (LSTM, $T{=}10$k, MUFL, $\omega{=}0.5$),
$-137.0\%$ (PatchTST, $T{=}10$k, MUFL, $\omega{=}0.7$), and
$-128.6\%$ (LSTM, $T{=}10$k, HUFL, $\omega{=}0.3$). The MUFL channel at
$T{=}10$k is particularly sensitive. TCN degrades consistently as well
($-7.6\%$ at $T{=}2$k, $-21.5\%$ at $T{=}10$k; $-14.5\%$ overall),
indicating that the
degradation is not architecture-specific and is not mitigated by $\Omega$
matching.

\subsection{Source Disagreement at Matched $\Omega$}
\label{sec:compare}

\begin{figure}[t]
\centering
\includegraphics[width=0.92\linewidth]{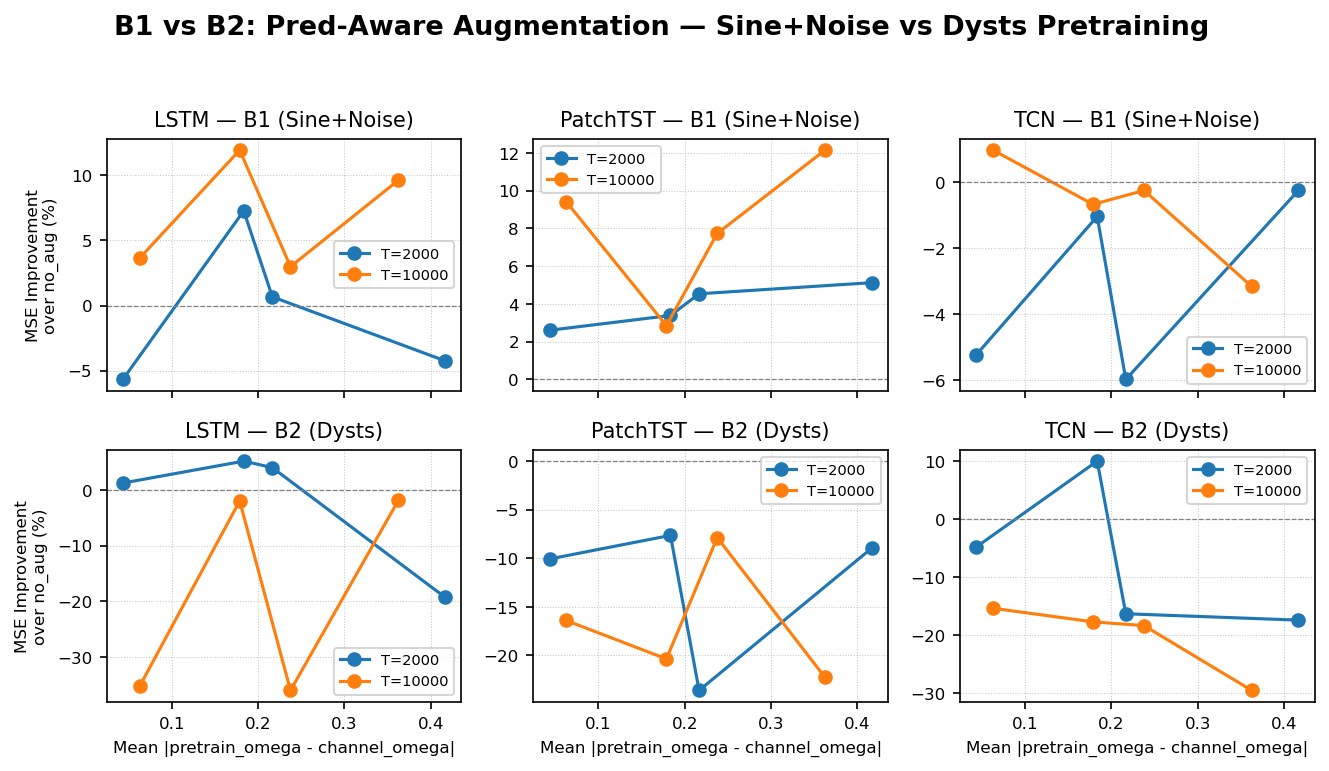}
\caption{Sine-plus-noise vs.\ dysts ODE per-configuration improvement at
  matched $\Omega$ ($2\times3$ grid over source/model). The near-random
  scatter ($r=-0.02$, $53\%$ directional agreement) shows $\Omega$
  matching alone does not determine transfer quality.}
\label{fig:b1_vs_b2}
\end{figure}

Fig.~\ref{fig:b1_vs_b2} compares the two sources at matched $\Omega$.
Were $\Omega$ the determining factor, the per-configuration outcomes would
align. Instead, they are nearly uncorrelated. Pooling all three
architectures ($168$ configurations), we correlate each configuration's
improvement under sine-plus-noise against its improvement under dysts ODE
at the \emph{same} target $\Omega$. This asks a different question from
Sec.~\ref{sec:b1}: not whether the $\Omega$ gap predicts one source's
outcome, but whether two $\Omega$-matched sources rise and fall together.
They do not: agreeing on the \emph{direction} of transfer (both improve or
both degrade) for only $53\%$ of configurations (barely above chance) with
Pearson $r=-0.02$. Matching $\Omega$ therefore carries almost no
information about whether a given source will improve performance on real
data.

This divergence admits a structural explanation. Chaotic ODE trajectories
exhibit multi-scale oscillations, sharp amplitude transitions, and
sensitive dependence on initial conditions, none of which are present in ETTm1.
Even after global $\Omega$ matching, this local temporal structure remains
incompatible with the target, producing representational mismatch during
fine-tuning. Sine-plus-noise, being spectrally simple and locally smooth,
is structurally closer to quasi-periodic real-world load signals, and
consequently transfers more favorably despite identical $\Omega$.

\section{Discussion}
\label{sec:disc}

\pgraph{Interpretation.}
Under ideal conditions, $\Omega$ matching behaves as hypothesized and the
$\Omega$ gap accounts for most of the variation in transfer quality. On real data, it is not, as two
sources at the same target $\Omega$ diverge almost completely
($r=-0.02$). $\Omega$ matching can therefore help in controlled settings
but is \emph{not sufficient} to predict transfer across sources. In
practice, $\Omega$ is best employed as a coarse \emph{screen} (to exclude
sources with gross mismatch) rather than as a guarantee of beneficial
transfer. It is thus most dependable for interpolating within a single
spectrally simple generator, as in our synthetic-to-synthetic study, and
least dependable for selecting between generator \emph{families}.

\pgraph{Factors beyond $\Omega$.}
The dysts failure implicates properties that $\Omega$ does not capture,
as $\Omega$ is a single global summary of the PSD and is insensitive to
other signal characteristics. Three properties are plausible candidates.
(i)~\emph{Waveform smoothness and higher-order structure}: the sharp
amplitude transitions and cusps of the ODE sources have no analog in
ETTm1, whereas smooth sine-plus-noise transfers more effectively at
identical $\Omega$. (ii)~\emph{Correlation structure}: the autocorrelation
decay and cross-channel dependence of the source may need to resemble
those of the target, not merely its spectral entropy.
(iii)~\emph{Source geometry and phase continuity}: sensitive dependence on
initial conditions renders ODE trajectories locally unpredictable in ways
that a global spectral measure averages out. An effective selection
criterion would likely combine $\Omega$ with such \emph{local}
descriptors.

\pgraph{Guidance for practitioners.}
We recommend: (1)~preferring spectrally simple, locally smooth generators
(sine-plus-noise, spectral interpolation) over chaotic ODE libraries, which
match $\Omega$ but introduce structural artifacts; (2)~using $\Omega$ only
to screen gross mismatches, not to rank close candidates; and
(3)~validating a candidate source on a small held-out target prefix before
large-scale pretraining, since the sign of transfer is not predictable from
$\Omega$.

\begin{figure}[t]
\centering
\includegraphics[width=\linewidth]{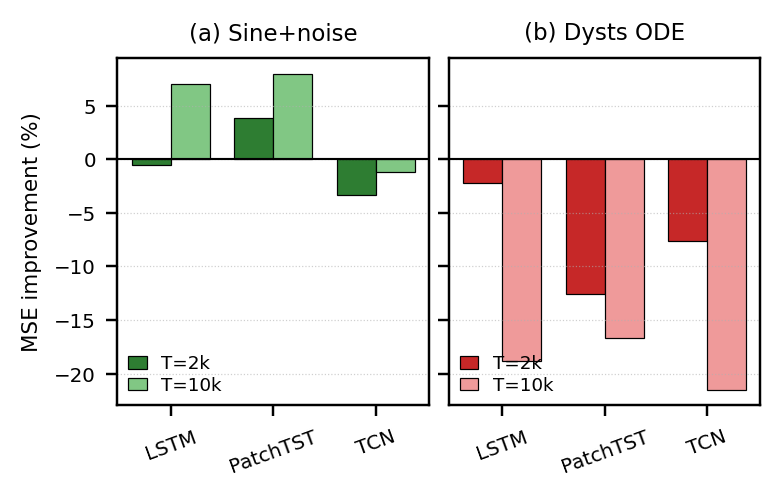}
\caption{Architecture-dependent response. Mean MSE improvement (\%) vs.\
  \texttt{no\_aug} by model and dataset size $T$ for each synthetic source.
  The three architectures show three distinct patterns: LSTM and PatchTST
  benefit from spectrally simple sine-plus-noise (a), TCN stays roughly
  neutral, yet all three degrade under $\Omega$-matched chaotic ODE data
  (b). Matching $\Omega$ does not equalize these outcomes.}
\label{fig:arch}
\end{figure}

\pgraph{Architecture-dependent asymmetry.}
The architectures respond differently to pretraining, and $\Omega$ does not
account for the difference. In the ideal setting
(Table~\ref{tab:directional}), the response depends on the \emph{direction}
of the $\Omega$ gap: LSTM gains from down-gap but is harmed by up-gap,
whereas PatchTST shows the opposite, indicating that recurrent and
attention-based models encode predictability differently. TCN forms a third
pattern, near-neutral in both directions ($-2.5\%$/$-1.7\%$). This carries
over to real data (Fig.~\ref{fig:arch}): TCN barely responds to
sine-plus-noise, whereas LSTM and PatchTST benefit, yet all three degrade
under chaotic ODE data. Three distinct response patterns at matched
$\Omega$ reinforce the central conclusion: the dominant factor is source
geometry and its interaction with the model, not $\Omega$.

\pgraph{Limitations.}
The evidence covers one real-world benchmark (ETTm1) and two synthetic source families, so generalization to other domains and generators remains open. The candidate factors are motivated by the dysts failure rather than isolated in a controlled ablation, and the reported correlations are descriptive rather than causal. Finally, the $\omega=0.9$ setting reflects achieved values of $\Omega\approx0.80$--$0.89$, while three seeds provide only a coarse estimate of run-to-run variance.

\pgraph{Conclusion.}
We asked whether spectral predictability $\Omega$ is a reliable criterion
for selecting synthetic pretraining data under distribution shift. The evidence suggests otherwise. $\Omega$ matching is effective under
ideal, single-source conditions but fails across sources: two generators
matched to the same target $\Omega$ transfer to ETTm1 in near-uncorrelated
ways ($r=-0.02$), and chaotic ODE data degrades every architecture. A
reliable criterion will likely need local waveform structure, correlation
geometry, and architecture-specific effects in addition to $\Omega$, and
chaotic ODE data should not be used as a pretraining surrogate without
prior validation of its compatibility with the target.

\clearpage
\section{Acknowledgments}
\label{sec:ackn}

\pgraph{Acknowledgments.}
We thank our colleagues and collaborators for their valuable 
feedback and support throughout this research. We thank the Software and Digital 
Platforms Department Management, specifically Richard Guinto, Johann Ignacio, and
Carlos Bating, for supporting this study. Analog Garage and Atulya Yellepeddi 
deserve special mention for their insightful discussions and contributions. 
Special thanks for the constructive feedback and guidance go to our technical 
reviewers on the Analog Devices Technical Review Board.

\pgraph{AI assistance.}
Generative AI tools were used to assist with drafting and revising the manuscript, 
coding, and LaTeX formatting. The research methodology, experimental design, scientific claims, 
interpretation of results, and conclusions were developed and validated by the authors, 
who take full responsibility for the final manuscript.

\bibliographystyle{IEEEbib}
\bibliography{strings,refs}

\end{document}